\documentclass[runningheads]{llncs}

\usepackage{amsmath,amssymb,amsfonts,booktabs,graphicx,mathtools}

\begin{document}
	
	
	\title{SynCGAN: Using learnable class specific priors to generate synthetic data for improving classifier performance on cytological images}
	\titlerunning{SynCGAN}
	\author{Soumyajyoti Dey\inst{1} \and
			Soham Das\inst{1} \and
	     	Swarnendu Ghosh\inst{1} \and
			Shyamali Mitra\inst{1} \and
			Sukanta Chakrabarty\inst{2} \and Nibaran Das\inst{1,*}}
		\authorrunning{S. Dey et al.}
		\institute{	Jadavpur University, Kolkata 700032, WB, India,
					\email*{nibaran.das@jadavpuruniversity.in} \and 
					Theism Medical Diagnostics Centre, Kolkata 700030, India,
					\email{drsukantachakraborty@gmail.com}}

	\maketitle
	
	\begin{abstract}
		One of the most challenging aspects of medical image analysis is the lack of a high quantity of annotated data. This makes it difficult for deep learning algorithms to perform well due to a lack of variations in the input space. While generative adversarial networks have shown promise in the field of synthetic data generation, but without a carefully designed prior the generation procedure can not be performed well. In the proposed approach we have demonstrated the use of automatically generated segmentation masks as learnable class-specific priors to guide a conditional GAN for the generation of patho-realistic samples for cytology image. We have observed that augmentation of data using the proposed pipeline called ``SynCGAN'' improves the performance of state of the art classifiers such as ResNet-152, DenseNet-161, Inception-V3 significantly.
		
	\end{abstract}

	
	\section{Introduction}

	The modern machine learning algorithms such as deep learning, have been greatly dependent on the availability of a large amount of high-quality data. But for various niche domains such as medical imaging large quantities of data are generally unavailable due to various constraints, such as lack of patients, infrastructural inadequacy, noisy environments, lack of experts for annotations and so on. However, with the advent of \textit{g}enerative \textit{a}dversarial \textit{n}etworks(GANs) \cite{goodfellow2014generative}, an avenue for high quality data generation has opened. In its base form, GANs are capable of generating samples from a randomly sampled prior which demonstrates likeliness to a predefined data distribution. However, without proper guidance, the generation process can result in eccentric outputs. However, \textit{c}onditional GANs (CGANs)\cite{mirza2014conditional}, on the other hand, use a semantically sensible prior for guiding the data generation process to generate more accurate and meaningful samples. In the proposed work, we explore the ability of CGANs to work with learnable priors for efficient data generation to improve classifier performance on cytology images. In most practical cases the number of available data samples is too limited for deep learning approaches to thrive. Thus data augmentation serves as a primary tool for improving learning ability. Though annotating pixel specific masks for cytology images is a difficult and expensive job, however, with adequate expertise and a decent amount of labor it is possible to annotate at least a small batch of samples for a better semantic representation. The proposed approach makes use of such semantic masks to serve as a prior for CGANs. While generating fully detailed cytology images without priors is much difficult, the generation of segmentation masks from scratch is a much simpler task given that the output distribution is binomial. Our proposed approach makes use of this factor to create learnable segmentation masks that can guide CGANs for synthetic data generation. Some relevant studies are discussed in the next section. The proposed methodology is provided in the subsequent section followed by experimental setups, results, and discussions in section 4 and future scopes are discussed in the conclusion thereafter. 
	
	\section{Related Works}
	Most common methods for data augmentation involved affine transformations\cite{ghosh2019reshaping} such as translation, rotation, scaling, shear, flipping and so on. Also, it has been noticed that training with added noise results in a much more robust classifier. The introduction of \textit{g}enerative \textit{a}dversarial \textit{n}etworks (GANs) has brought a shift in the paradigm of generative processes in computer vision. Several approaches have made use of GANs for data augmentation. Adar et al. \cite{frid2018gan} proposed a GAN based liver lesion data augmentation technique where after the extraction of ROI for classification was done using CNN.  Dataset was augmented in two ways: i) the ROIs were augmented by affine transformations ii) the synthetic data was generated from ROIs using DCGAN(Deep Convolution Generative adversarial Network) and ACGAN(Auxiliary Classifier GAN). DCGAN showed greater performance compared to ACGAN. Shin, et al.\cite{shin2018medical} proposed a GAN based model to segment tumor of brain MRI images of two traditional datasets: ADNI and BRATs. Normal brain MRI images were segmented using an image to image translation model using CGAN\cite{isola2017image}. Synthetic abnormal brain MRI scans were obtained from labels(tumors) by incorporating some changes in the label(e.g. increasing the size, changing the position of the label, or placing the tumor in a healthy brain MRI segmentation map). The synthetic images were used in data augmentation for training the model. Improved performance of tumor segmentation was observed by adding the synthetic data to the real data but without using normal data augmentation methods. Tom et al.\cite{tom2018simulating} simulated patho-realistic ultrasound images of the IVUS dataset using deep generative models. Tissue echogenicity maps were generated from the ground truth of the dataset. From these maps simulation of ultrasound images was produced using a physics-based simulator. Two-staged GAN was used to generate patho-realistic ultrasound images and stability of training state. In the first stage, images from the simulator were taken as input to GAN from which low-resolution images were generated. In the second stage, these low-resolution images from the first stage of GAN were transformed into high-resolution images. Bissoto et al.\cite{bissoto2018skin} suggested a GAN based model to generate high-resolution images of skin lesion of the ISIC challenge dataset. Classifiers were trained on real data as well as on synthetic data.

	\section{Proposed Methodology}
	
	The goal of the current work is to generate realistic cytology images similar to images collected during FNAC(Fine Needle Aspiration Cytology) test. The cytology images were collected from Theism Medical Diagnostics Centre, Dumdum, Kolkata. These cytological data were mainly collected by FNAC test, and were captured using an Olympus microscope at 40X magnification in the presence of the professional practitioners. Around 156 cytology images were collected among which 77 were benign samples and 79 were malignant samples.

	\begin{figure}[htbp]
		\centering
		\includegraphics[width=0.8\textwidth]{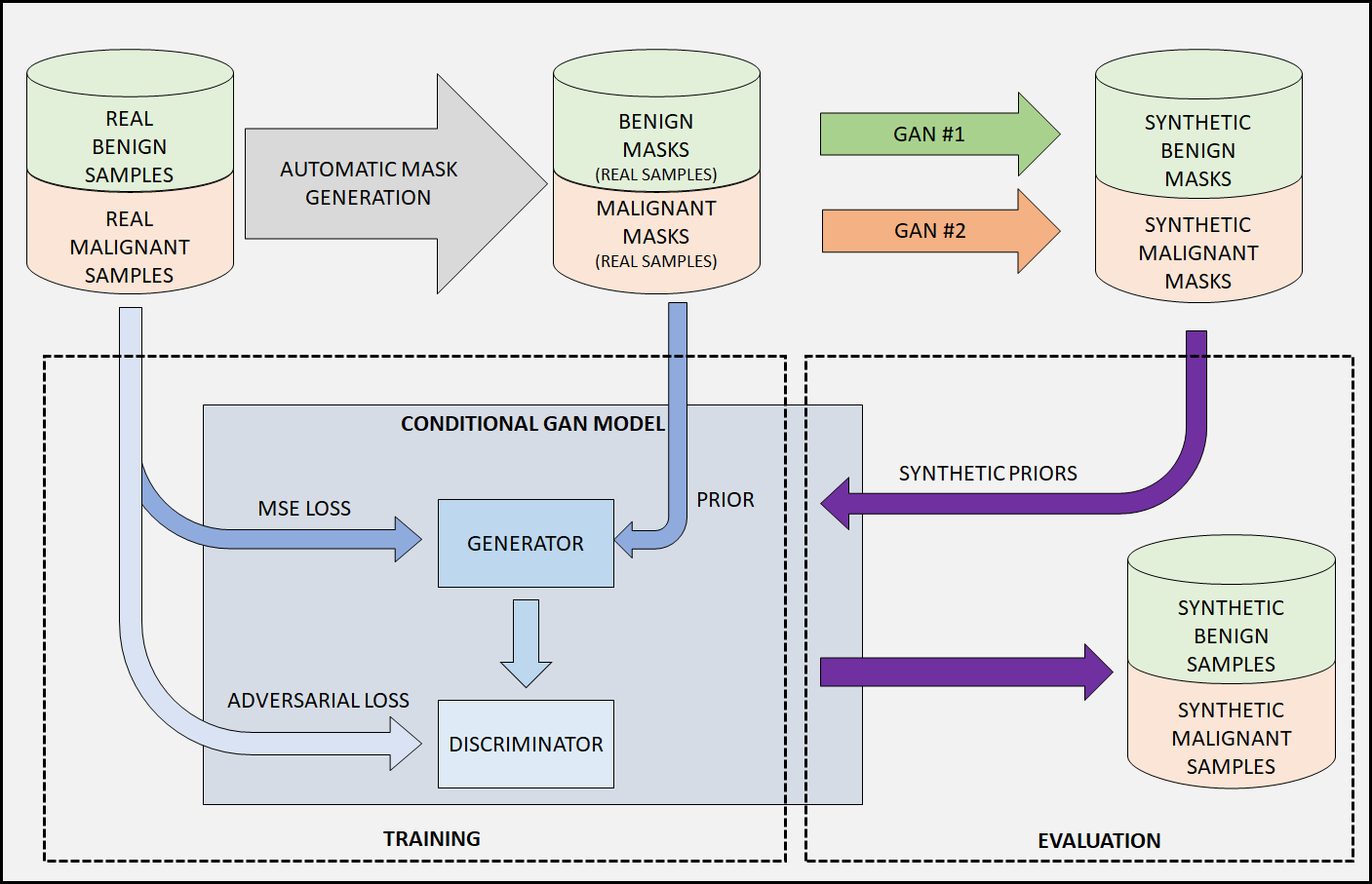}
		\caption{SynCGAN: The proposed data augmentation pipeline}
	\end{figure}

	\subsection{The data augmentation pipeline}
	\label{sec:pipeline}
	The proposed method of synthetic data generation consists of 3 phases. Firstly, segmentation masks are collected from real images using an unsupervised method. Secondly, a CGAN is trained on these pairs of images and auto-generated segmentation masks. Two sets of synthetic segmentation masks are generated using a GAN for each of the two classes. Finally, these synthetic segmentation masks are used to guide the previously trained CGANs to generate patho-realistic samples for data augmentation. 
	
	\subsection{Mask generation}
	\label{sec:mask}
	The proposed methodology requires a set of pixel-level annotations to guide a CGAN for data generation. Due to the lack of hand-annotated samples, an unsupervised approach was used for nuclei segmentation. There have been many developments in the field of image segmentation lately\cite{Ghosh2019UnderstandingDL}. For our work, first, the contrast of the RGB cytology image is increased by the histogram stretching method. The image is then converted to a grayscale image. To eliminate the irrelevant portions, adaptive thresholding \cite{kowal2011computer} based segmentation algorithm is adopted and the RGB image is converted to a binary segmented mask. But the red blood cell, cytoplasm which had similar high local contrast is distinguished using the Gaussian Mixture clustering algorithm. Finally, the refined binary segmented mask is extracted. The presence of an unsupervised mask generation technique alleviates the necessity of large amount of training data.
	
	\begin{figure}[htbp]
		\centering
		\includegraphics[width=0.8\textwidth]{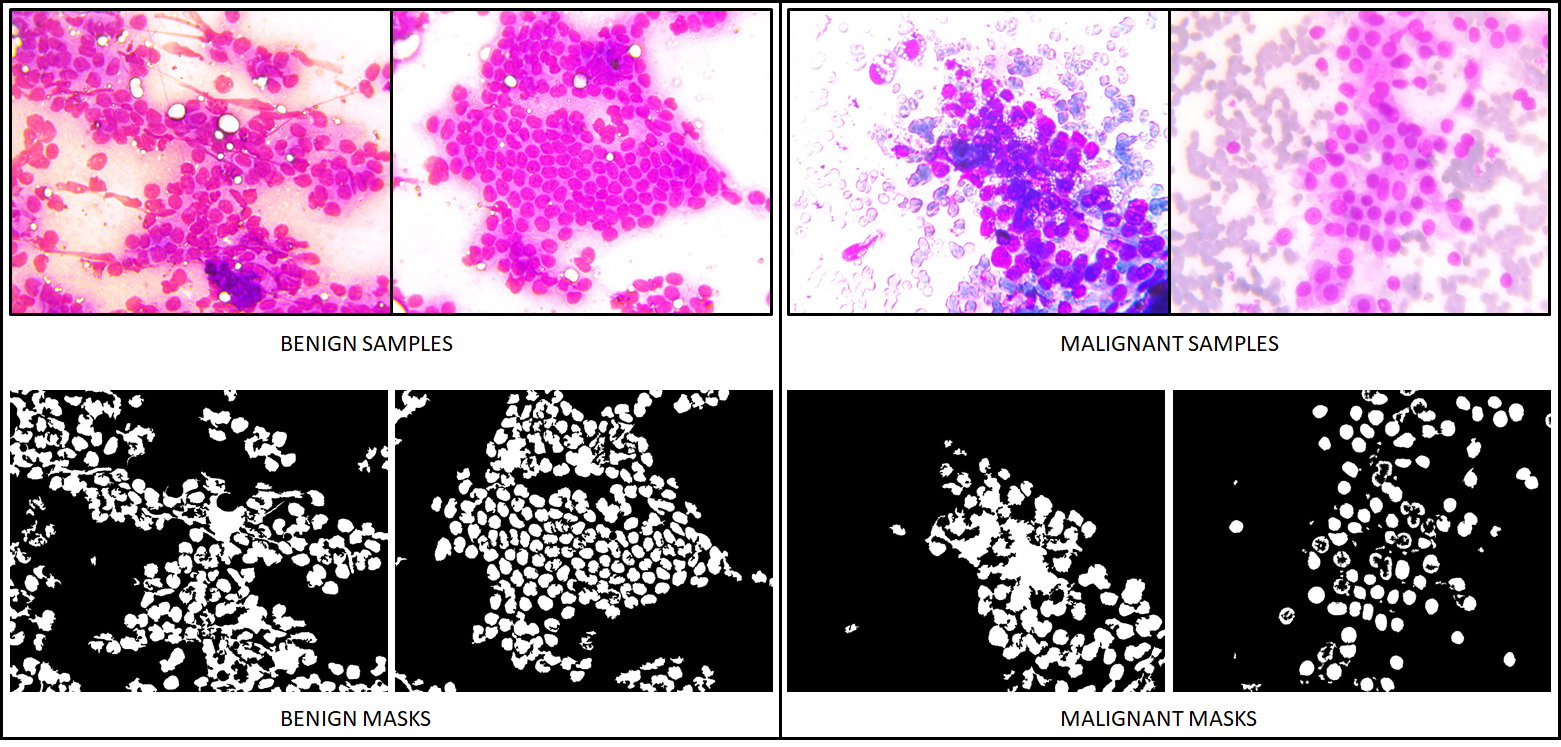}
		\caption{Auto generated segmentation masks using \cite{kowal2011computer} from real samples}
	\end{figure}
	
	\subsection{Training the CGAN to generate patho-realistic RGB images from segmentation masks}
	\label{sec:CGAN}
	Generating RGB images from scratch using a traditional GAN is difficult as the generation procedure can be represented as a prediction of 256-dimensional multinomial distribution across three channels for each pixel. However, the segmentation mask is simply a pixel specific binomial distribution which is much easier to predict when starting without a predefined prior. Thus a CGAN~\cite{mirza2014conditional} must be trained which takes the segmentation masks generated in the previous step as a prior and a generator loss is reduced against the corresponding RGB image. We derive inspiration from the pix2pix network ~\cite{isola2017image}. For the generator, we use a modified UNet like architecture. Normally the UNet architecture used transposed convolution for upscaling the feature maps. However, that results in checkerboard artifacts due to overlap of kernels during the fractional stride. Instead, bilinear interpolation opted for upscaling the feature map followed by a $3\times3$ convolution layer for refinement. The discriminator network has been directly implemented using the PatchGAN discriminator as demonstrated in ~\cite{isola2017image}. The discriminator attempts to detect real and fake samples from the dataset and the generator respectively. The objective function $V$ can be written as:
	\begin{align}
	\min_{G} \max_{D} V(D,G)  = \mathbb{E}_{(x,y)\sim p_{data}(x,y)}[\log(D(x|y))] \nonumber \\
	+ \mathbb{E}_{y \sim p_{data}(y)}[\log(1-D(G(y)))] .
	\end{align}
	
	Here $G$ and $D$ refers to the generator and discriminator. $x$ represents the RGB sample, $y$ represents the corresponding auxiliary representation which serves as a prior for the generator. The $x$ and $y$ samples are drawn from the input data distribution $p_{data}(x,y)$ that consists of RGB images and their auxiliary representations or segmentation masks in the current scenario. In our case, the auxiliary representations are the automatically generated segmentation masks as described in section \ref{sec:mask}.
	
	\par At every iteration, the discriminator and the generator are trained alternatively as was performed previously. During the training of the discriminator, the segmentation mask and its corresponding RGB image are concatenated on its channel dimension. It is then passed through the discriminator (Patch GAN)\cite{isola2017image} and discriminator loss of the real image is calculated as below
	
	\begin{equation}
	D_{loss} = -\log(D(x))+\log(1-D(G(y))),
	\end{equation}
	
	where $x$ represent samples from the input database and $y$ refers to segmentation masks of those samples. A binary cross-entropy loss function is used to calculate the adversarial loss.
	While training the generator, the segmentation masks are passed through the generator network and the loss is calculated. The loss has two components denoting the adversarial loss exhibited by the discriminator and the mean square error between the generated sample and the actual RGB image from the dataset that corresponds to the mask $y$.
	
	\begin{equation}
	G_{loss} = - \log(D(G(y)))+\lambda\ \text{MSE}(G(y),x|y)
	\end{equation}
	
	where $\lambda$ is the weight of Mean Squared Error (MSE) loss. The weight of the mean squared error loss is set to $100$ based on empirical analysis on a small validation set. $x$ and $y$ represent samples from the RGB image and the segmentation mask dataset.

	\subsection{Training the GAN to generate segmentation masks}
	\label{sec:GAN}
	A conditional GAN(CGAN) usually generates synthetic samples conditioned by some predefined priors. In the current scenario, the CGAN has been trained to generate samples from segmentation masks highlighting the spatial distribution of nuclei across the cytoplasm. To generate patho-realistic synthetic samples during the evaluation phase a class-specific prior distribution is necessary. For that purpose, we train a GAN model ~\cite{goodfellow2014generative} to generate binary segmentation masks based on a randomly drawn seed from a gaussian distribution. While models like CycleGAN~\cite{zhu2017unpaired} can be used for image translation, it is not suitable for synthetic data generation. The most straight forward method to generate synthetic samples would have been to train an end-to-end GAN. However, it has been noticed in ablation studies that without a prior the quality of outputs is very poor. The primary reason being the complexity of predicting the intensity value of a pixel. Given the output image has three channels, each pixel exists within a search space with $256^3$. However, a binary segmentation mask is a much easier output to predict given that each pixel belongs to a binomial distribution. On the other hand, the shape information encoded within these segmentation masks is quite informative about the class of the samples, namely, benign or malignant. We train two separate GANs trained on segmentation masks belonging to each of the predefined classes.
	
	The objective of the GAN network is simply defined as 
	\begin{align}
	\min_{G} \max_{D} V(D,G)  =\nonumber 
	\mathbb{E}_{x\sim p_{data}(x)} [\log(D(x))] \\+ \mathbb{E}_{z \sim p(z)}[\log(1-D(G(z)))]
	\end{align}
	
	Here $x$ refers to samples drawn from the input data distribution $p_{data}(x)$. $z$ refers to randomly sampled priors from a Gaussian distribution $p(z)$. $G$ and $D$ refers to the generator and discriminator network. The architecture of the GAN used in the current work is very similar to the one described in the previous section. It consists of a generator inspired from UNet whos transposed convolutions have been replaced with bilinear interpolation for upscaling and a convolution layer for refinement. The discriminator is derived from the PatchGAN discriminator as demonstrated in ~\cite{isola2017image}. During the training phase, the discriminator loss is given by,
	\begin{equation}
	D_{Loss} = -(\log(D(x))+\log(1-D(G(z))))
	\end{equation}
	
	and the generator loss is given by:
	\begin{equation}
	G_{Loss} = - \log(D(G(z)))  
	\end{equation}
	
	However, due to very low number of samples, the discriminator was too overpowering and saturates at a very early stage. To deal with this issue some additional measures were taken as described below\cite{kumar2019c}. 
	\begin{itemize}
		\item \textbf{Label smoothing:} 
		The labels for real and fake samples are set as 1 and 0 by default. To enforce some fuzziness in the system, a random number between 0.9 and 1 was taken for real samples and a random number between 0.1 and 0 was taken for fake samples while training. However, this is unnecessary while training the generator as we want to bottleneck the learning curve of the discriminator and not the generator.
		
		\item \textbf{Randomly flipping labels:} To even further confuse the discriminator at some random iterations real samples are labeled as fake and vice versa. This confusion results provide some breathing space for the generator so that it can learn the requisite features.    
	\end{itemize}
	\par Other models such as Wasserstein GAN\cite{arjovsky2017wasserstein} can further improve results.
	
	\subsection{Evaluating on the trained model to generate synthetic images}
	The final phase generates class specific patho-realistic synthetic samples. According to the pipeline discussed in section \ref{sec:pipeline}, at first the class-specific GANs are used to generated segmentation masks (refer section \ref{sec:GAN}). Then the generated segmentation masks are fed into the trained CGAN model (refer section\ref{sec:CGAN}) to obtain the RGB synthetic images. The synthetic data distribution is modeled as,
	\begin{equation}
	p_{data(z)} = \mathbb{E}_{z \sim p(z)}[G_{CGAN}(G_{GAN}(z))] 
	\end{equation}
	where, $G_{CGAN}$ and $G_{GAN}$ refers to the generators of the CGAN and GAN model described in section \ref{sec:CGAN} and \ref{sec:GAN} respectively.
	\par While a simple CGAN trained on a prior denoting the class of samples could also be viable, having a richer representation like a mask provides more flexibility in terms of variations in the synthetic dataset.
	
	\section{Experiments and Results}
	
	The main objective of the work is to generate class-specific synthetic data similar to microscopic cytology images that can boost the performance of standard image classification algorithms. For our experiments we have used three classifiers namely, ResNet-152\cite{he2016deep}, Inception-V3\cite{szegedy2016rethinking} and DenseNet-161\cite{huang2017densely}. Each of these networks has a proven track record in tough image classifications tasks such as the ILSVRC. In the original dataset, there were 156 images in total. Out of which 77 were benign samples and 79 were malignant samples. For the benefit of a cleaner calculation, a total of 150 images were selected with 75 images from each class. The dataset was randomly divided in the ratio of 3:1:1 into training, validation, and testing set with an equal number of samples in each class. The ratio of synthetic data to original training data was maintained at 2:1. Thus 90 synthetic training images were generated per class.
	\begin{figure}[htbp]
		\centering
		\includegraphics[width=0.75\textwidth]{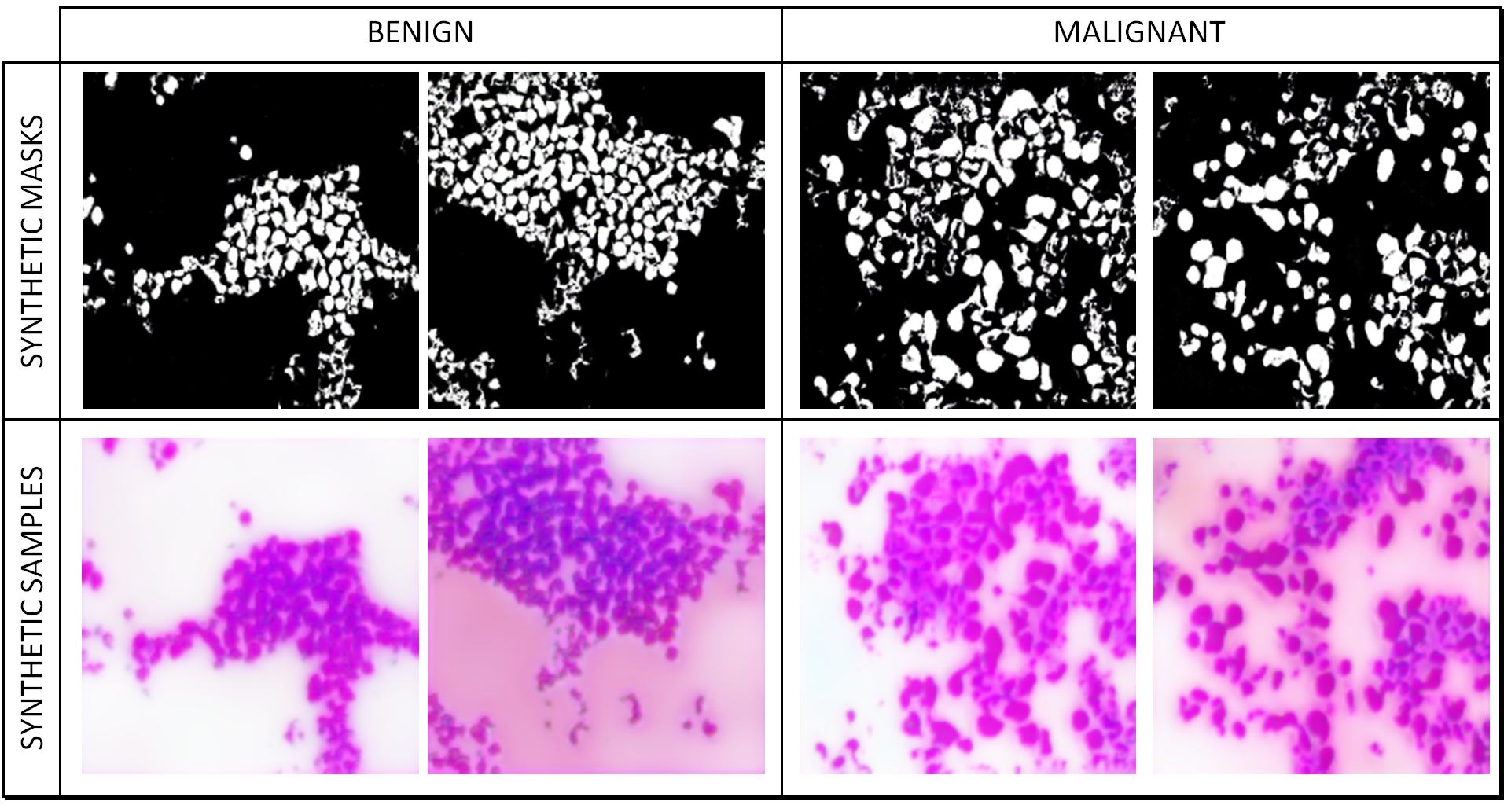}
		\caption{Synthetic masks generated by class specific GANs (top) are fed as priors to the conditional GAN during evaluation phase to generate patho-realistic samples (bottom) for benign(left) and malignant(right) classes separately}
	\end{figure}
	\begin{table}[]
		\centering
		\caption{Number of samples in the original and synthetic dataset}
		\label{tab:dataset}
		\begin{tabular}{@{}llll@{}}
			\toprule
			\textbf{Dataset}          & \textbf{Training} & \textbf{Validation} & \textbf{Testing} \\ \midrule
			\textit{Original Images}  & 90                & 30                  & 30               \\
			\textit{Synthetic Images} & 180               & -                   & -                \\ \bottomrule
		\end{tabular}
	\end{table}

	\subsection{Experimental Setup}
	The performance metric defining the goodness of the synthetic data generation pipeline, referred to as SynCGAN in the current work, is given by its impact on the test accuracy obtained using the three previously mentioned network, namely, ResNet-152, Inception-V3, and DenseNet-161.
	The experiment was conducted to analyze the impact of synthetic data augmentation on several grounds. 
	\begin{enumerate}
		\item Impact of augmentation using data generated by SynCGAN,
		\item Performance of SynCGAN generated data augmentation against traditional data augmentation,
		\item Performance of SynCGAN generated data augmentation against GAN generated data augmentation.
	\end{enumerate}
	
	The CGAN model trained for a maximum of 200 epochs and the best model was saved based on minimum generator loss on the validation dataset. While the GAN model was trained for almost until the generator loss saturated(1600 epochs). For both the cases adam optimizer was used. All the experiments were conducted on Nvidia GTX 1060 GPU.
	
	\subsection{Observations and Analysis}
	The first observation as shown in table \ref{tab:results_1}, shows that augmentation of data generated with the proposed SynCGAN improves the performance of classifiers. When compared with traditional augmentation techniques like random horizontal and vertical flipping, random rotation and addition of Gaussian noise, the proposed method of augmentation has a higher impact. When traditional data augmentation was combined with SynCGAN based augmentation, the performance was either at par or lower than exclusive SynCGAN based augmentation.
	\begin{table}[]
		\centering
		\caption{Performance of classifiers while using the dataset with and without augmentation. Orig: Original Data, Prop: Data generated using the proposed SynCGAN pipeline, Trad: Data generated using traditional augmentation techniques}
		\label{tab:results_1}
		\begin{tabular}{@{}lcccc@{}}
			\toprule
			\textbf{Classifier}   & \textbf{ Orig } & \textbf{ Orig+Prop } & \textbf{ Orig+Trad } & \textbf{ Orig+Trad+Prop } \\ \midrule
			\textit{ResNet-152}   & 73.33         & \textbf{76.67}     & 70.00              & \textbf{76.67}          \\
			\textit{DenseNet-161} & 80.00         & \textbf{86.67}     & 83.33              & 84.67                   \\
			\textit{Inception-V3} & 73.33         & \textbf{80.00}     & 63.33              & 76.67                   \\ \bottomrule
		\end{tabular}
	\end{table}

	Secondly, as a control to our proposed model, we implemented a purely GAN based pipeline for data augmentation, which performed far below the proposed model as shown in table \ref{tab:results_2}. This GAN based architecture also had a generator and discriminator network similar to our proposed model for a fair comparison.
	\begin{table}[]
		\centering
		\caption{Performance of classifiers while using the dataset with and without augmentation. Orig : Original Data, Prop: Data generated using proposed SynCGAN pipeline, GAN: Data generated using vanilla GAN.}
		\label{tab:results_2}
		\begin{tabular}{@{}lccccc@{}}
			\toprule
			\textbf{Classifier}            & \textbf{ Orig }& \textbf{ Prop } & \textbf{ Orig+Prop } & \textbf{ GAN } & \textbf{ Orig+GAN } \\ \midrule
			\textit{\textbf{ResNet-152}}   & 73.33           & 73.33           & \textbf{76.67}       & 50.00          & 63.33  \\
			\textit{\textbf{DenseNet-161}} & 80.00          & 63.33           & \textbf{86.67}       & 50.00          & 60.00  \\
			\textit{\textbf{Inception-V3}} & 73.33           & 66.67           & \textbf{80.00}       & 56.67          & 66.67  \\ \bottomrule
		\end{tabular}
	\end{table}

	\section{Conclusion}
	In the present work, a CGAN based data augmentation technique has been proposed using class specific priors that improves the performance of various state of the art CNNs such as ResNet-152, DenseNet-161, and Inception-V3 on cytology images corresponding to FNAC tests. Unlike a normal GAN, we have used learnable segmentation masks as class-specific priors to guide a conditional GAN for more robust synthetic data generation. It is to be noted that the method is quite dependent on the mask generation algorithm and hence extensive studies may be performed using other available nuclei segmentation techniques for further analysis. Furthermore, the method may be further generalized to adapt to other types of cytological data.
	
	\section*{Acknowledgement}
	This work is funded by SERB (DST), Govt. of India sponsored project ( order no. EEQ/2018/000963 dated 22/03/2019). The authors are thankful to Theism Medical Diagnostics Centre, Kolkata, West Bengal, India for providing cytology samples and also thanks to Centre for Microprocessor Application for Training, Education, and Research, Jadavpur University for providing additional infrastructure for the research.
	\bibliographystyle{splncs04}
	\bibliography{ref}

\begin{thebibliography}{10}
\providecommand{\url}[1]{\texttt{#1}}
\providecommand{\urlprefix}{URL }
\providecommand{\doi}[1]{https://doi.org/#1}

\bibitem{bissoto2018skin}
Bissoto, A., Perez, F., Valle, E., Avila, S.: Skin lesion synthesis with
  generative adversarial networks. In: OR 2.0 Context-Aware Operating Theaters,
  Computer Assisted Robotic Endoscopy, Clinical Image-Based Procedures, and
  Skin Image Analysis, pp. 294--302. Springer (2018)

\bibitem{frid2018gan}
Frid-Adar, M., Diamant, I., Klang, E., Amitai, M., Goldberger, J., Greenspan,
  H.: Gan-based synthetic medical image augmentation for increased cnn
  performance in liver lesion classification. Neurocomputing  \textbf{321},
  321--331 (2018)

\bibitem{goodfellow2014generative}
Goodfellow, I., Pouget-Abadie, J., Mirza, M., Xu, B., Warde-Farley, D., Ozair,
  S., Courville, A., Bengio, Y.: Generative adversarial nets. In: Advances in
  neural information processing systems. pp. 2672--2680 (2014)

\bibitem{he2016deep}
He, K., Zhang, X., Ren, S., Sun, J.: Deep residual learning for image
  recognition. In: Proceedings of the IEEE conference on computer vision and
  pattern recognition. pp. 770--778 (2016)

\bibitem{huang2017densely}
Huang, G., Liu, Z., Van Der~Maaten, L., Weinberger, K.Q.: Densely connected
  convolutional networks. In: Proceedings of the IEEE conference on computer
  vision and pattern recognition. pp. 4700--4708 (2017)

\bibitem{isola2017image}
Isola, P., Zhu, J.Y., Zhou, T., Efros, A.A.: Image-to-image translation with
  conditional adversarial networks. In: Proceedings of the IEEE conference on
  computer vision and pattern recognition. pp. 1125--1134 (2017)

\bibitem{kowal2011computer}
Kowal, M., Filipczuk, P., Obuchowicz, A., Korbicz, J.: Computer-aided diagnosis
  of breast cancer using gaussian mixture cytological image segmentation.
  Journal of medical informatics \& Technologies  \textbf{17} (2011)

\bibitem{mirza2014conditional}
Mirza, M., Osindero, S.: Conditional generative adversarial nets. arXiv
  preprint arXiv:1411.1784  (2014)

\bibitem{shin2018medical}
Shin, H.C., Tenenholtz, N.A., Rogers, J.K., Schwarz, C.G., Senjem, M.L.,
  Gunter, J.L., Andriole, K.P., Michalski, M.: Medical image synthesis for data
  augmentation and anonymization using generative adversarial networks. In:
  International Workshop on Simulation and Synthesis in Medical Imaging. pp.
  1--11. Springer (2018)

\bibitem{szegedy2016rethinking}
Szegedy, C., Vanhoucke, V., Ioffe, S., Shlens, J., Wojna, Z.: Rethinking the
  inception architecture for computer vision. In: Proceedings of the IEEE
  conference on computer vision and pattern recognition. pp. 2818--2826 (2016)

\bibitem{tom2018simulating}
Tom, F., Sheet, D.: Simulating patho-realistic ultrasound images using deep
  generative networks with adversarial learning. In: 2018 IEEE 15th
  International Symposium on Biomedical Imaging (ISBI 2018). pp. 1174--1177.
  IEEE (2018)

\end{thebibliography}
	
\end{document}